\documentclass{ecai} 
\usepackage{subfigure}
\usepackage{latexsym}
\usepackage{amssymb}
\usepackage{amsmath}
\usepackage{amsthm}
\usepackage{booktabs}
\usepackage{enumitem}
\usepackage{graphicx}
\usepackage{color}
\usepackage{algorithm}
\usepackage{algorithmic}
\newcommand{\our}{\texttt{Logical-GLM}}
\newcommand{\joey}[1]{{\color{black} #1 \color{black}}}
\newcommand{\jinkb}[1]{{\color{black} #1 \color{black}}}

\newcommand{\zf}[1]{{\color{black} #1 \color{black}}}
\newcommand{\base}[1]{{\color{black} #1 \color{black}}}
\newcommand{\ignore}[1]{{}}
\newcommand{\BibTeX}{B\kern-.05em{\sc i\kern-.025em b}\kern-.08em\TeX}

\begin{document}

%%%%%%%%%%%%%%%%%%%%%%%%%%%%%%%%%%%%%%%%%%%%%%%%%%%%%%%%%%%%%%%%%%%%%%%%

\begin{frontmatter}

%%% Use this command to specify your submission number.
%%% In doubleblind mode, it will be printed on the first page.
%\paperid{328} 
%% Self-defined macros

%%% Use this command to specify the title of your paper.

\title{Planning with Logical Graph-based Language Model \\for Instruction Generation
}

%%% Use this combinations of commands to specify all authors of your 
%%% paper. Use \fnms{} and \snm{} to indicate everyone's first names 
%%% and surname. This will help the publisher with indexing the 
%%% proceedings. Please use a reasonable approximation in case your 
%%% name does not neatly split into "first names" and "surname".
%%% Specifying your ORCID digital identifier is optional. 
%%% Use the \thanks{} command to indicate one or more corresponding 
%%% authors and their email address(es). If so desired, you can specify
%%% author contributions using the \footnote{} command.

\author[A]{\fnms{Fan}~\snm{Zhang}
%\orcid{....-....-....-....}%
\footnote{Email: zhangf356@mail2.sysu.edu.cn}
}
\author[B]{\fnms{Kebing}~\snm{Jin}%\orcid{....-....-....-....}
\footnote{Email: kbjin@gzu.edu.cn}
}
\author[A]{\fnms{Hankz Hankui}~\snm{Zhuo}
%\orcid{....-....-....-....}
\thanks{Corresponding Author. Email: zhuohank@gmail.com}} 

\address[A]{School of Computer Science and Engineering, Sun Yat-sen University, Guangzhou, China
}
\address[B]{Department of Computer Science, Guizhou University \\ \vspace{1em}To appear at ECAI 2024}
%\address[C]{Short Alternate Affiliation of Third Author}

%%% Use this environment to include an abstract of your paper.

\begin{abstract}
Despite the superior performance of large language models to generate natural language texts, it is hard to generate texts with correct logic according to a given task, \joey{due to the difficulties for neural models to capture strict logic from free-form texts. In this paper, we propose a novel graph-based language model, {\our}, to extract strict logic from free-form texts and then infuse into language models.} Specifically, we first capture information from natural language instructions and construct logical probability graphs that generally describe domains. Next, we generate logical skeletons to guide language model training, infusing domain knowledge into language models. At last, we alternately optimize the searching policy of graphs and language models until convergence. The experimental results show that {\our} is both effective and efficient compared with traditional language models, despite using smaller-scale training data and fewer parameters. Our approach can generate instructional texts with more correct logic owing to the internalized domain knowledge. Moreover, the search of logical graphs reflects the inner mechanism of the language models, which improves the interpretability of black-box models.
\end{abstract}

\end{frontmatter}

%%%%%%%%%%%%%%%%%%%%%%%%%%%%%%%%%%%%%%%%%%%%%%%%%%%%%%%%%%%%%%%%%%%%%%%%

\section{Introduction}
\joey{
Instruction generation aims to generate a course of actions to complete the given task. It is an open issue which requires both sophisticated and knowledge-intensive reasoning abilities of models \cite{francis2022core,wang2023lana}. %这里加个引用！！！
Inspired by the reasoning capabilities of large language models (LLMs), works have been committed to investigate how LLMs guide the instruction generation in specific domains \cite{kim2024language,liang2023code}. However, the LLMs can internalize only the general world knowledge rather than domain-specific logic, which is required for instruction generation. As mentioned by Valmeekam et al. \cite{valmeekam2022large}, even in a seemingly simple environment like BlocksWorld, LLMs are quite ineffective in generating reasonable instructions. 
% can still not generate reasonable instructions.
%\cite{weidinger2021ethical,kassner2020pretrained}

An interesting instance is utilizing LLMs to generate instructional texts to guide robots in cooking \cite{recipe}. 
% As mentioned by Katserelis et al., instructional texts generated by language models are generally inconsistent and often impractical in the real world \cite{recipe}.
In detail, an example generated recipe for task "frog legs with lemon and thyme" is: \textit{In a pan* Gently melt the lemon; Add the oil; A little salt and a pinch of sugar; Bring to the boil and then cover and cook over a medium heat for 25 minutes; Stirring occasionally; Until the frogs are soft; Remove from the heat; Leave to cool slightly and stir in the remaining lemon juice; Place the loaf of bread on a work surface sprinkled with flour; Sprinkle with a little flour and quickly knead together to make a smooth; }

% Even ball of dough; Beat the egg whites until stiff; Then fold into the mixture with the cheese; Roll out the dough to a rectangle measuring 30 x 40 cm; Put in a baking dish lined with baking parchment; Cover and brush the surface lightly with non-stick baking paper; Bake for 10-15 minutes until golden.}

The above example shows that the generated recipe contains many actions whose preconditions are not satisfied, which hinders robots from completing the action sequence in the real world. For instance, a robot can execute \textit{"In a pan* Gently melt the lemon"} only if \textit{"puts lemon into the pan"} has been carried out. 
% is executed in advance, the robot can commit \textit{"In a pan* Gently melt the lemon"} successfully. 
Next, the above example generates incomplete instructions, e.g., lacking corresponding operation objects. For example, the second sentence, \textit{"Add the oil"} should be completed as \textit{"Add the oil into the pan"}. Furthermore, the generated actions gradually deviate from the given task. In the above example, all of the actions from the ninth sentence to the end seem to be relevant to "bake a cake" rather than the given task. 
% CoT, ToT

To augment the reasoning abilities of LLMs for generating reasonable instructions, works have been proposed to leverage planning thought to ensure strict logic. Wei et al. \cite{wei2022chain} generate a chain of thought (CoT) to improve the ability of LLMs to perform complex reasoning. Similarly, Yao et al. \cite{yao2024tree} leverage tree of thought (ToT) to perform deliberate decision making by considering multiple different reasoning paths and self-evaluating choices to decide the next course of action, simultaneously looking ahead or backtracking when necessary to make global choices. Both CoT and ToT transfer the planning thought into LLM generation to ensure the strict logic. The strict logic require the support of knowledge. Though LLMs internalize general world knowledge and behave well on a range of common-sense tasks, they behave poorly on domain-specific tasks, which require domain knowledge. 
% Common-sense knowledge of LLMs contains rare domain knowledge and even domain knowledge often differs from common-sense, e.g., in the industrial environment. 
\jinkb{Reality applications, e.g. industrial environment, require specific domain knowledge, which may not be included in the training common-sense dataset.}Inspired by this, we are curious if it is possible to import domain knowledge for supporting AI Planning in domain-specific generation. By doing this, we aim to answer the question: \textbf{how domain knowledge can be better mixed with LLM knowledge for supporting AI Planning}. 

Indeed, there have been attempts to infuse domain knowledge into LLMs \cite{liu2024drak,wang2021hierarchical,ciatto2024large}.} For example, Wang et al. \cite{wang2021hierarchical} propose a novel approach named Hierarchical Concept-Driven Language Model (HCDML) to infuse knowledge into language models. HCDML learns a Hierarchical Language Model for implementing planning thought and a Recurrent Conceptualization-enhanced Gamma Belief Network for internalizing domain logic. However, the internalized logic is not strict since the training data can contain redundant or even wrong texts and hinder HCDML from supporting planning thought when generating texts. This indicates extracting strict logic from texts is difficult. To do this, we represent domain knowledge in the form of symbolic logic, namely PDDL (Planning Domain Description Language) \cite{aeronautiques1998pddl}. Then, we construct a logical graph to internalize the strict logic of extracted structural expressions.
% In AI Planning, structural expressions can remove such unimportant information and remain core logic of texts. As done by AI planning, \jinkb{we represent domain knowledge in the form of symbolic logic, namely PDDL (Planning Domain Description Language) \cite{aeronautiques1998pddl}.} Then, we construct a logical graph to internalize the strict logic of extracted structural expressions. To embed strict logic into language models, we present an EM-style framework where language models and constructed logical graphs guide each other for decision-making and text generation. 

The core steps of our {\our} are as follows. Firstly, we convert natural language instructions into action sequences in the form of PDDL, so as to eliminate redundant and incorrect information. Actions in classical planning consist of preconditions and effects, where the interlaced predicates form complex logical relations in the domain. Therefore, we consider that the sequential order of actions implies the domain logic, which potentially reflects logical constraints and updating between actions. Thus, we then filter and leverage converted instructions to extract action pairs \joey{with higher frequency than the threshold we set and construct logical probability graphs using structuralized action pairs. Meanwhile, we train language models using converted instructional texts. Finally, the logical graph and language model guide each other to make final decisions. Specifically, in the search process, the probability graph generates logical skeletons towards the target task node under the guidance of LMs. The target-oriented search can ensure the actions always related to the task. And the precedence order of action nodes ensure the preconditions of actions can be satisfied. }

\begin{figure}[!ht]
    \centering
    %\vspace{-4mm}
    \includegraphics[width=0.5\textwidth]{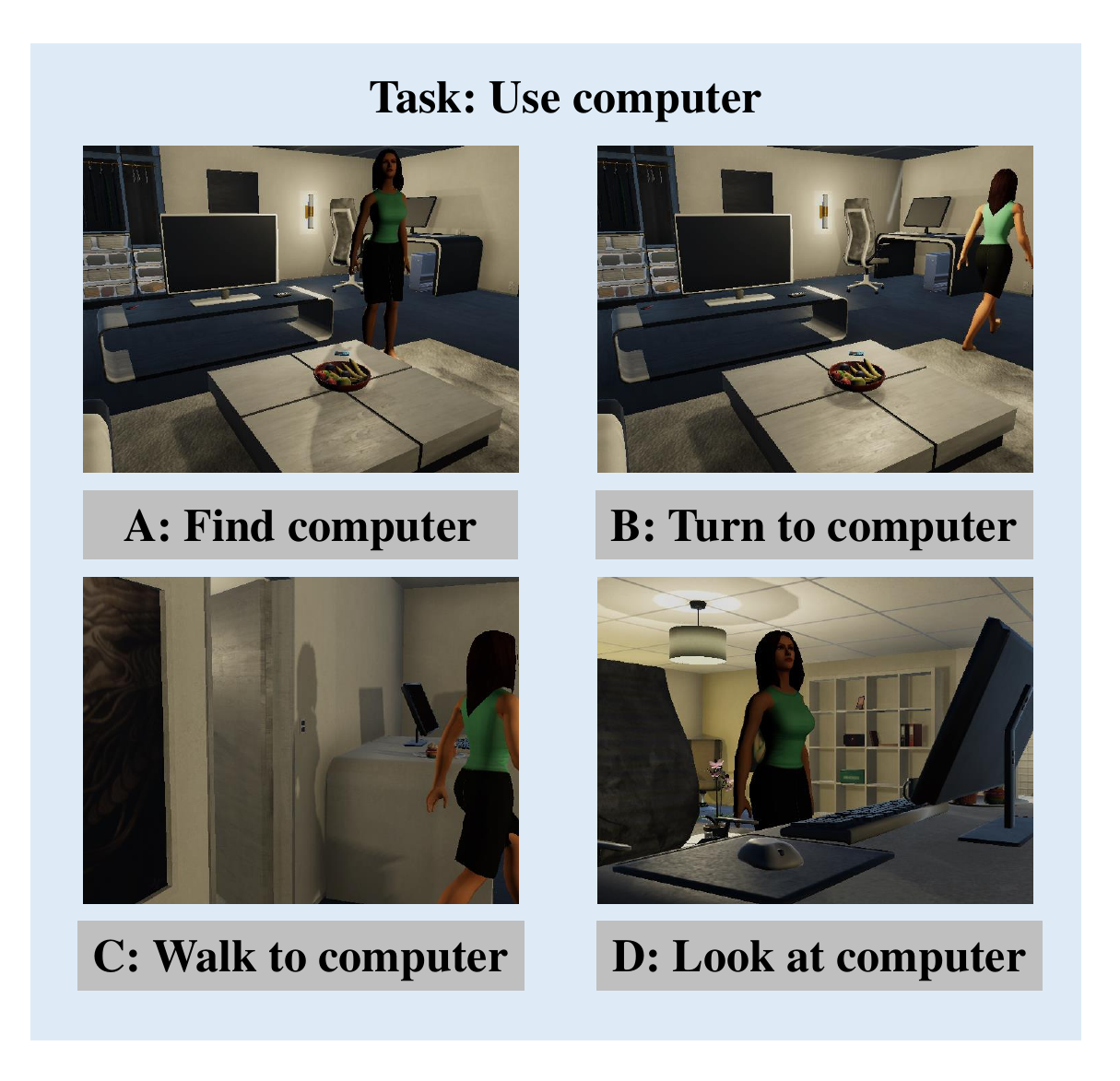}
    %\vspace{-4mm}
    \caption{Sample action generated by our {\our} and execution in VirtualHome}
    \label{render}
\end{figure}

\section{Problem Definition}
\label{cheap:1}
Let $T_x$ denote the task for completion, where \emph{executable} and \emph{correct} instructional texts are expected to be generated (i.e., Figure \ref{render}). The label texts and generated texts are denoted by $\hat{T_y}$ and $T_y$, respectively. In this way, our training data can be defined as a series of tuples $\Phi$, where each tuple consists of $\langle T_x,\hat{T_y}\rangle$, where $T_x$ and $\hat{T_y}$ denote the task and corresponding label instruction texts. Given a task that does not exist in training data as input, the goal of the proposed model is to generate simpler and more logical instructional texts with a smaller amount of parameters compared to large language models. Specifically, the generated instructional texts should be executable and correct to ensure the completion of the given task. We will discuss the detailed evaluation metrics in the experimental section.

\section{Logical Graph-based Language Model}
\label{model}
% node, edge, core words, objects, adjanceny
When generating instructional texts by large language models, there are several major challenges: (1) The forms of texts are variant and different texts may express the same domain logic. (2) Texts contain some redundant information, which influences logic extraction. (3) The amount of actions and objects is large, which adds difficulties to logic extraction. (4) Gradient explosion can happen for fine-tuning large-scale language models. To tackle these challenges, we propose our {\our} which includes:
% \begin{enumerate}
%     \item A logical graph that learns core logic from action traces of each task in a specific domain;
%     \item A language generation model which both learns logic from the constructed logical graph and generalizes well to unseen tasks;
%     \item An EM-style framework where logical graph and language models guide each other to internalize domain logic
% \end{enumerate}

\quad 1. A logical graph that learns core logic from action traces of each task in a specific domain;

\quad 2. A language generation model that both learns logic from the constructed logical graph and generalizes well to unseen tasks;

\quad 3. An EM-style framework where logical graphs and language models guide each other to internalize domain logic.

\begin{figure}[htp]
    \centering
    \includegraphics[width=0.5\textwidth]{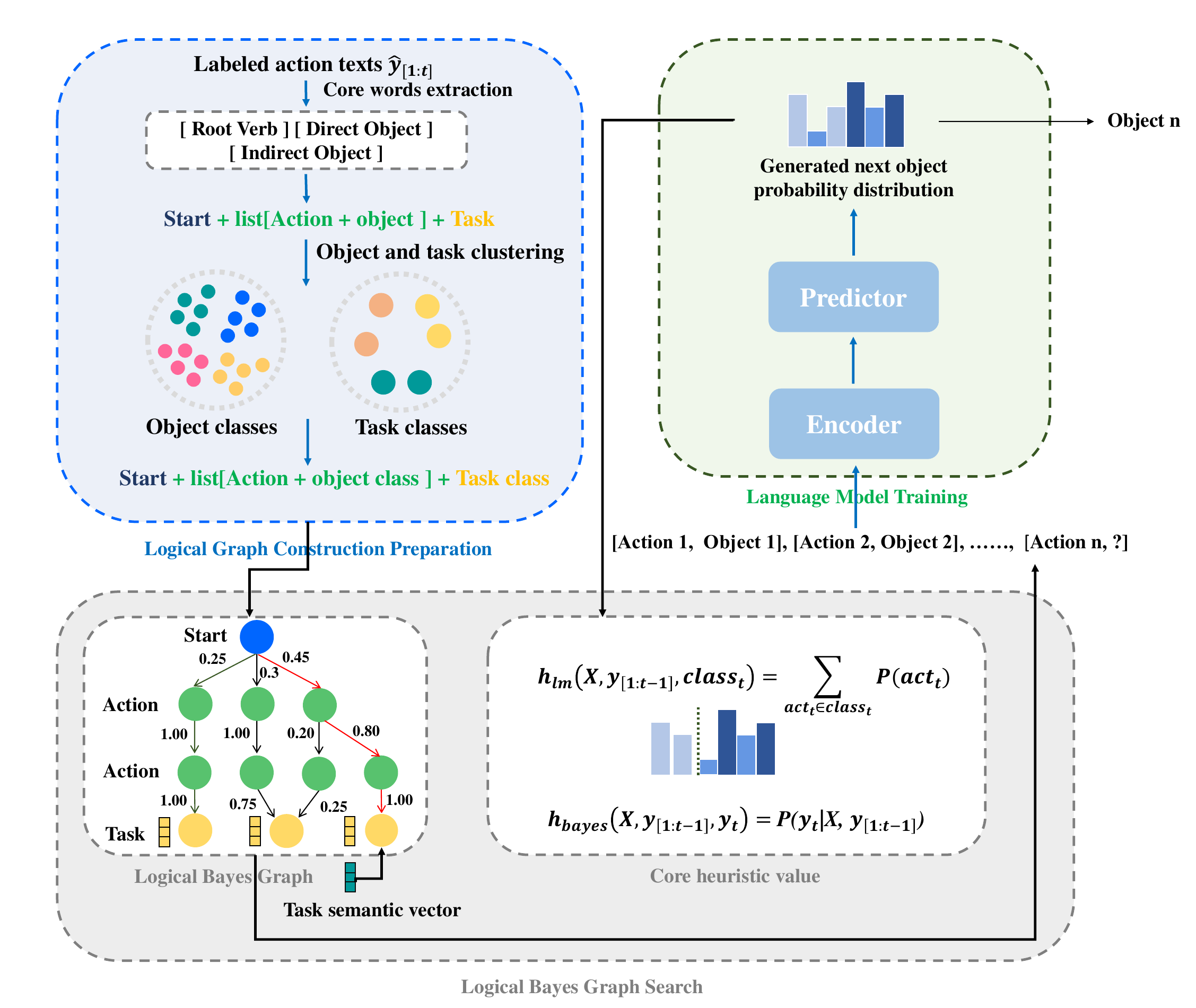}
    \caption{An overview of the proposed {\our}}
    \label{alg}
    %\vspace{-4mm}
\end{figure}

The overall algorithm is presented in Figure \ref{alg}. Our model is composed of following major components: Logical Graph Construction Preparation (blue part), Logical Porbability Graph Search (gray part); Language Model Training and Logical Graph Optimization (green part).
\subsection{Domain-specific logical graph construction preparations}
To construct logical graph, we design a preprocessing module, which refines sophisticated logical relations in natural language instructions (upper left part of Figure \ref{alg}). \joey{In detail, the logical relation between action A and action B means that action B can only be executed when action A has been committed. }

If we directly adopt natural language instructions to mine rules, redundant and incorrect information may affect relation extraction. Therefore, for extracting refined logical relations, we first use semantic parser spaCy\footnote{https://spacy.io/} to extract the core verbs and objects in the domain. \joey{Then, we leverage pre-trained RoBERTa model to calculate the semantic vectors of extracted objects and tasks \cite{reimers-2019-sentence-bert}. After that, we }use K-means to cluster extracted objects and tasks according to the the semantic vectors. An object detector then substitutes objects with clustered object class. Since policy selection considers multiple previously executed actions in reinforcement learning, we then analyze relations between previous $k (k\in Z, k \ge 1)$ actions with current action for extracting logical relations. The frequency of action pair within k time steps is calculated and compared with the threshold $\epsilon (\epsilon \in R, 0 \le \epsilon \le 1)$. \zf{The threshold $\epsilon$ depends on the specific action (Equation \ref{preconditional}). For each action, we calculate the frequency mean of previous actions plus the unbiased estimation of frequency variance within k time steps.}The action pairs with higher frequency than $\epsilon$ are retained. In this way, we finally acquire refined logical rules for constructing logical graphs.
\joey{
\begin{equation}
\label{preconditional}
% \mathcal{H}=\\ \arg\max_{a \in a_{type}} h_{dis} - h_{local} * 0.3 - h_{global} * 0.7 - h_{lm} * 0.1
\epsilon=\bar{p}+\frac{\sum_{i=1}^t\left(p_i-\bar{p}\right)^2}{t-1}, t\in\mathbb{Z}, p \in \mathbb{R}
\end{equation}
}
\zf{where $t$ is the number of previous adjacent actions of the current action. $p_i$ denotes the frequency of each previous action and $\bar{p}$ is the average frequency of all previous actions.} \\
% To express the relation rules implied in natural language instructions, we first construct a logical graph according to the events and arguments. However, redundant and incorrect information included in texts results in difficulties to generate reliable logical graphs. On the other hand, the vast quantities of objects make it hard to extract refined logical relations. To handle those challenges, we use language models to select words that representatively indicate sentences, extract arguments, and cluster them with different types. Then we borrow the syntax of action models used in AI planning to structuralize instructions with the selected words and learned types of objects. At last, we construct a logical graph based on inference of the relations between instructions.
\subsection{Logical probability graph construction}
%从头到尾重新进行描述
After acquiring refined logical rules, we then construct a logical probability graph to indicate the logical relations implied in the text. However, the natural language logical rules may contain much redundant information, so we first structuralize refined logical rules using syntax from AI planning \cite{DBLP:conf/aaai/ZhuoNK13, DBLP:journals/ai/ZhuoK17,DBLP:journals/ai/JinZXWK22}. Consistent with Planning Domain Definition Language (PDDL), \joey{we define the lifted action node as $action(?x-type)$ to reserve core verbs and objects, where $action$ denotes core verb and $?x-type$ denotes the object type without instantiation \cite{aeronautiques1998pddl}.} The lifted action node can be visited and then grounded into $action(instance)$. \joey{For example, $grab(?x-fruit)$ can be grounded as specific $grab(apple)$.} Following the same source node, we add lifted action nodes and task type node of each action sequence in order. If the path from source node to the current node exists, we just add next node following the current node. In this way, we internalize the domain logic into the graph skeleton. 
Motivated by the wide use of bayes networks, we endow each edge with transition probability to learn causal relationships \cite{heckerman2008tutorial}. In detail, we represent the transition probability using the corresponding frequency of action pair in training data. In this sense, we construct a logical probability graph which contains refined domain logic. 

\subsection{Logical graph search and heuristic value computation}
\joey{The language model has better generalization abilities than traditional planning \cite{bettergpt}.} However, even large language models only work on simple reasoning tasks and fail to solve sophisticated reasoning problems \cite{valmeekam2022large}.

Therefore, we design a logical probability graph search module, which generates \joey{instructional skeletons under the guidance of both logical probability graphs and language models.} To guide the search across the constructed logical graph, we design a heuristic value calculation module. The total heuristic value is calculated based on four factors:
\joey{
\begin{equation}
\label{totalheu}
% \mathcal{H}=\\ \arg\max_{a \in a_{type}} h_{dis} - h_{local} * 0.3 - h_{global} * 0.7 - h_{lm} * 0.1
\mathcal{H}=\\ - h_{dis}(.) - \omega_1h_{len}(.) + \omega_2h_{bayes}(.) + \omega_3h_{lm}(.)
\end{equation}
}
where $\mathcal{H}$ stands for the total heuristic value. $h_{dis}$, $h_{len}$, $h_{bayes}$ and $h_{lm}$ stand for distance, expected program length, bayes, and language model heuristic values. We define the program length as the number of sentences in instructional texts.  $\omega_1$, $\omega_2$ and $\omega_3$ are positive hyper-parameters. The detailed heuristic value calculation is formulated as follows \joey{and the candidate node with the highest heuristics can be selected}:
\joey{
\begin{equation}
\label{dis}
%% 函数？没有输入，前文用的是h_dis,统一！
h_{dis}(y_n, x)= 
\left\{ 
    \begin{array}{lc}
        0, & if\ \text{path\_exist}(y_n, x)\\
        \infty, & \text{otherwise}\\
    \end{array}
\right.
\end{equation}
}
where $h_{dis}$ implies whether the current node can reach the target task type node. $y_n$ denotes action type node at current step and $x$ the target task type node, respectively. $\text{path\_exist}(y_{n}, x)$ denotes whether exists one path from the current node to the target task type node. \zf{If there exists no path between $y_n$ and $x$ in the logical graph, this means the current action node is probably irrelevant to the task. When searching the graph, we give priority to nodes which can reach the \joey{target} task node.}
\textbf{\begin{small}
\begin{equation}
h_{len}\left(y_{[1:n]}, x, \hat{l}\right)=\frac{min(|avg\_len(y_n, x)+n-1-\hat{l}|, \hat{l})}{\hat{l}}
\end{equation}
\end{small}}
where $y_{[1: n]}$ denotes the previously $n-1$ selected action type nodes concatenated a current candidate action type node. $y_n$ denotes the current candidate node. $\hat{l}$ stands for the expected number of instructional sentences. \joey{$avg\_len(.)$ calculates the average length of all paths from the current candidate node to the target node. We use current steps plus $avg\_len(.)$ as the estimated length of our generated action sequence. Then, we calculate the difference between our estimated length and the expected length. The shorter the difference between our estimated instructional length and the ideal length, the larger the value of $ - \omega_1h_{len}(.)$. In this way, the length of our instructions can be adjusted according to the expected length.}
% \begin{small}
% \begin{equation}
% h_{bayes}\left(y_{[1:t]}^{type}, x^{tar}\right)=\frac{P\left(y^{type}_{[1: t]} \rightarrow x^{tar}\right) P\left(y_1^{type} \rightarrow y_t^{type}\right)}{P\left(y_1^{type} \rightarrow x^{tar}\right)}
% \end{equation}
% \end{small}
\joey{
\begin{equation}
h_{\text {bayes }}\left(y_{i}\mid y_[1: n-1], x\right)=\frac{P\left(y_{i}\rightarrow x\right)}{\sum_{y_{j} \in \text { sons }\left(y_{n-1}\right)} P\left(y_{j} \rightarrow x\right)}
\end{equation}
where $\rightarrow$ stands for the probability sum of all possible paths from the visited path to the target node. $sons$ denote the action nodes which can be visited from the last action node. Based on the already visited action nodes and target type nodes, we calculate the score of one action node among all candidates. 
}
\begin{equation}
\label{lmh}
h_{lm}(\overline{y}_n|x, \overline{y}_{[1:n-1]}) = \max_{\overline{y}_n \in grd(y_n)} P(\overline{y}_n|x^{tar}, \overline{y}_{[1:n-1]})\\
\end{equation}
where $\overline{y}_n$ stands for all possible grounded actions at the current step. \joey{For instance, $grab(?x-fruit)$ can be grounded into $grab(mango)$, $grab(bananas)$ and other possible specific actions.} $\overline{y}_{[1:n-1]}$ stands for grounded action node sequences at previous $n-1$ time steps. $h_{lm}$ calculates the probability of generating one possible grounded action given the texts of task and all previously grounded actions. The detailed probability calculation for each grounded action by the LM can be seen in Equation (\ref{pp}).\\
% As illustrated in Figure \ref{heu}, the graph search algorithm moves towards the task goal and the heuristic value guides the node selection. 
%这一段出大问题！！！
%怎么都说不清楚，啧啧
% \joey{
% \begin{equation}
% \label{pp}
% \mathrm{P}(\mathbf{grd})=\mathrm{P}\left(\operatorname{grd}_1\right) \prod_{i=2}^n P\left(\operatorname{grd}_i \mid \operatorname{grd}_1, \ldots, grd_{i-1}\right)
% \end{equation}

\joey{
\begin{equation}
\label{pp}
P=\gamma\left(T_x, T_{\bar{y}_{[1: n-1]}};T_{y_n}\right)
\end{equation}
where $\gamma$ denotes the language model for generating the next action. In detail, the texts of the task $T_x$ and previous actions $T_{\bar{y}_{[1: n-1]}}$ serve as the input, and we acquire the probability of LM generating a grounded action $T_{y_n}$.}
% The probability of LM generating one grounded action is the continued product of outputting each token given task, grounded actions and the previous generated tokens.
%}

\begin{figure}[!ht]
    \centering
    \includegraphics[width=0.5\textwidth]{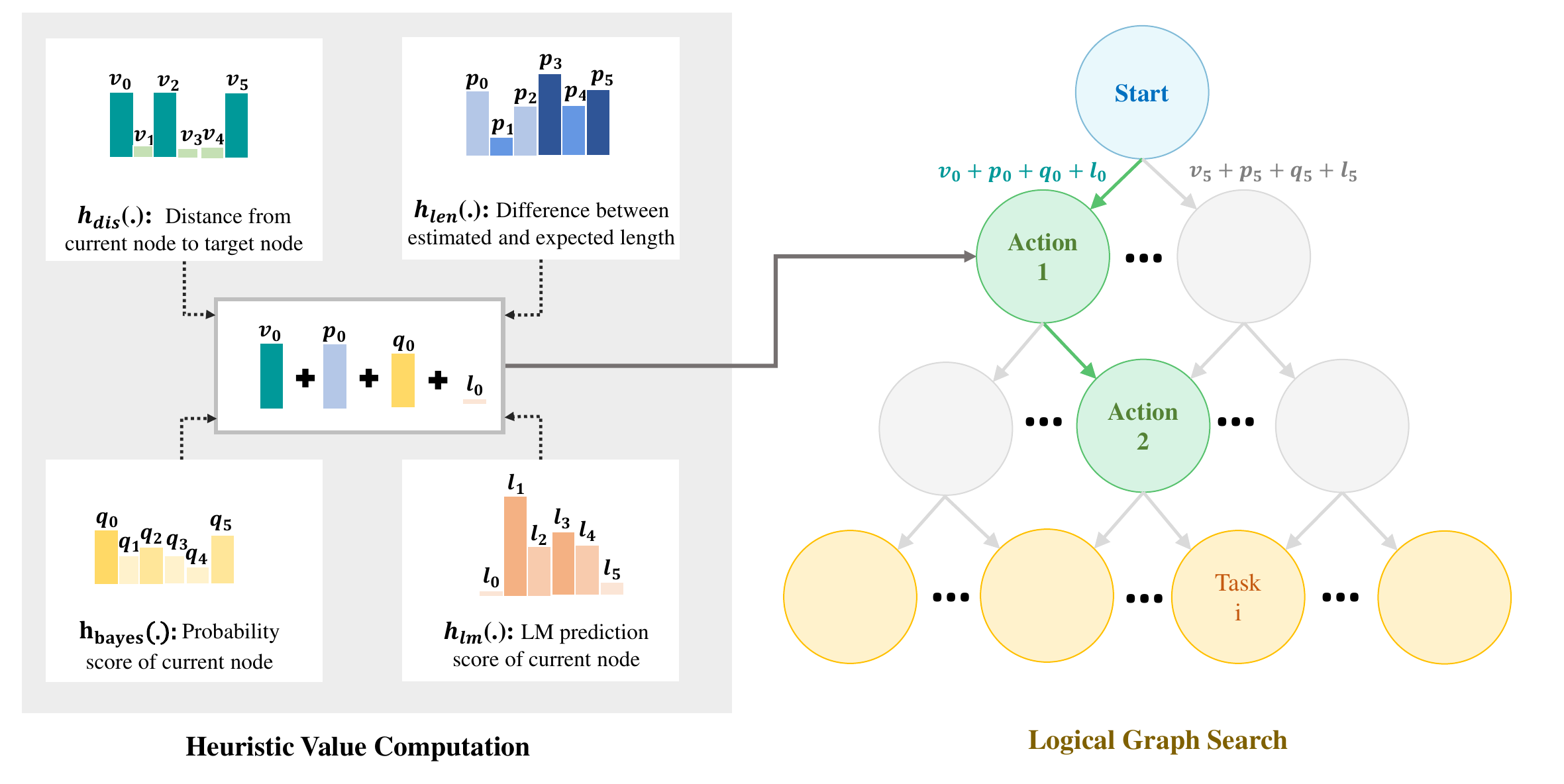}
    \caption{An example of domain-specific logical graph search}
%% 两边大小差距太多，看起来有点不好看
    \label{heu}
\end{figure}

We then use heuristic values to guide the logical graph search. As illustrated in Figure \ref{heu}, the graph search algorithm moves towards the task goal and the heuristic value guides the node selection. In detail, we first select the target task type node (yellow nodes) which has the highest semantic similarity with the current task text. Then, we start from the source node (blue node) and set it as the current node. At each step, we calculate the heuristic values of all successor nodes from the current node. \joey{For instance, the heuristic value of Action 1 is $v_0+p_0+q_0+l_0$, which corresponds to four heuristics defined in Equation (\ref{totalheu}).} The nodes with the highest heuristic values (green nodes) are selected as the current nodes and grounded into $action(instance)$. The node selection continues until reach the target task type node. Finally, we concatenate all grounded actions and output the generated instructional texts.
% 我还没解释ground的意思，这部分写完了需要在Logical bayes graph construction模块中予以解释

% calculates the probability of generating one possible grounded action given task and all previously grounded actions

% Specifically, the heuristic values contain distance, program length, bayes, and language model heuristic values. 

\subsection{Language model training and logical graph optimization}
The generated instructional texts from logical graphs internalize domain logic. Together with label instructions, we thus use generated instructional texts to guide language model training. The trained language model also guides the logical graph search. In this sense, logical graphs and language models guide each other to learn domain logic and make policies. \\
Specifically, each input to the language model is comprised of task description $x$ and label instructional text $\hat{y}$. Since both label instructions and generated instructional texts contain domain logic, we define the cross-entropy loss function $\mathcal{L}(\theta)$ as follows:
\begin{equation}
\label{loss}
\mathcal{L}(\theta)=-\mathcal{H}\left(f(y_{[1: t]}^{type}), f(y_{[1: t]})\right)-\mathcal{H}\left(f(\hat{y}_{[1: t]}), f(y_{[1: t]})\right)
\end{equation}
where $\mathcal{L}(.)$ is the loss function. $\mathcal{H}(.)$ stands for the cross entropy between two vectors. $f(.)$ is the embedding function. $\hat{y}_{[1:t]}$ is the label instructions and $y_{[1:t]}$ is the generated texts from language model. We use the above loss function to update parameters by stochastic gradient descent (SGD) \cite{boyd2004convex}. In this sense, the parameters of language model are updated to make the language model internalize logic contained in both label instructions and generated instructional texts from the logical graph. The trained language model then guides the logical graph search. In this way, language model and logical graph guide each other and better internalize domain logic.
\subsection{Overview of our {\our}}
The overall algorithm is presented in Algorithm \ref{alg:algorithm}. We first initialize parameters $\theta$ of the language model randomly (Line 1). After that, we use the processed annotated task tuples $\Phi=<x,\hat{y}>$ to construct a logical graph (Line 2). We then search the logical graph to generate instructional texts under the guidance of the language models (Line 5). After that, we update parameters $\theta$ of the language model by optimizing $\mathcal{L}(\theta)$ (Line 6). We repeat the above procedure until the stop requirement is satisfied and output the language model parameters $\theta$.

\begin{algorithm}[!ht]
\caption{EM-style Optimization Algorithm}
\label{alg:algorithm}
\textbf{Input}: Annotated task tuples $\Phi=<x,\hat{y}>$\\
% \textbf{Parameter}: \\
\textbf{Output}: Logical Probability Graph and well-trained language model
\begin{algorithmic}[1] %[1] enables line numbers
\STATE Initialize parameters $\theta$ of the language model randomly;
\STATE Construct the logical graph using annotated task tuples $\Phi=<x,\hat{y}>$;
\FOR{iteration $\leftarrow$ 1...R}
%\WHILE{condition}
\STATE Randomly Pick annotated tasks $<x,\hat{y}>$;
% \IF {conditional}
\STATE Generate instructional texts by logical graph search with language model guidance (Equation (\ref{heu}));
% \ELSE
\STATE Update $\theta$ using both annotated task tuples and generated instructional texts (Equation (\ref{loss}));\\
% \ENDIF
%\ENDWHILE
\ENDFOR
\STATE \textbf{return $\theta$}
\end{algorithmic}
\end{algorithm}

\section{Experiments}
\label{exp}
\subsection{Datasets}
To evaluate the effectiveness of the proposed model, we investigate the textual reasonability, executability and correctness of generated texts. To evaluate the executability, we need an environment where textual interactions are supported. Most existing embodied environments except VirtualHome are unsuitable. We thus choose VirtualHome as our environment \cite{puig2018virtualhome}.

\begin{figure}[!ht]
    \centering
    \includegraphics[width=0.5\textwidth]{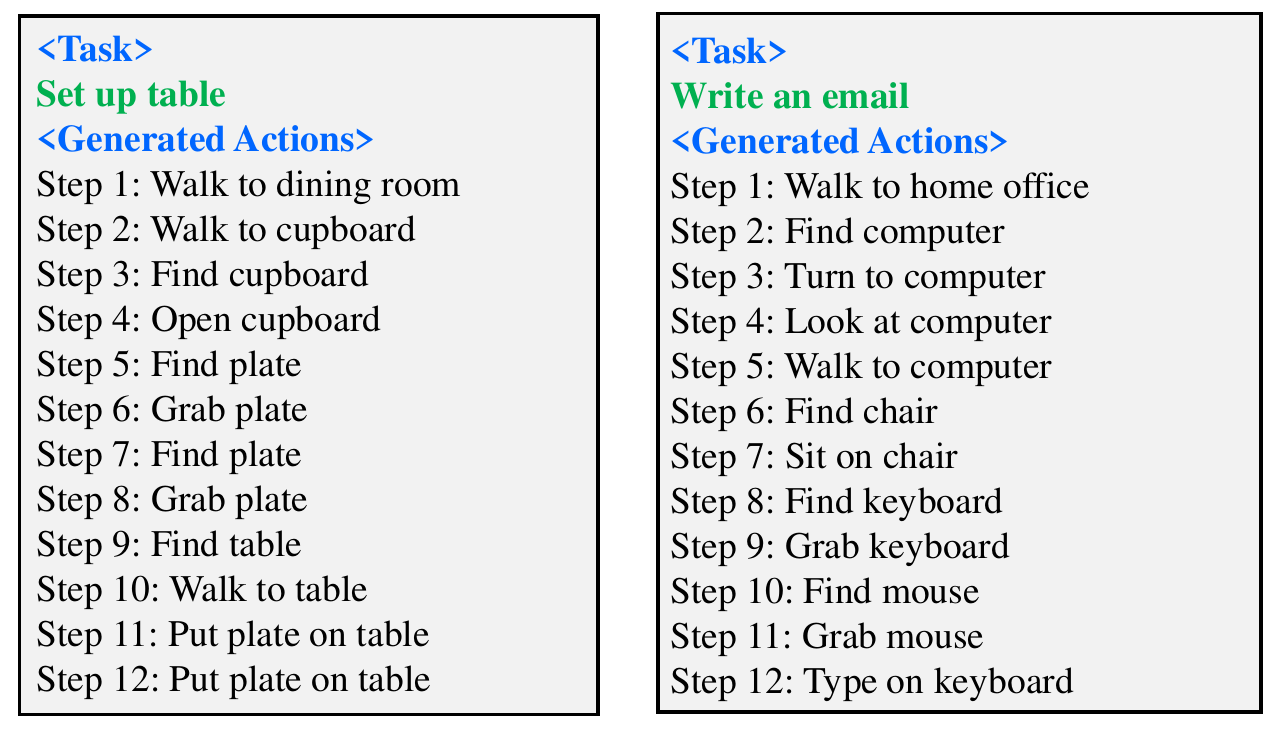}
    \caption{An example of training data}
%% 两边大小差距太多，看起来有点不好看
    \label{demo}
\end{figure}

In total, there are 5048 pieces of available instructional texts for 202 different tasks. A vivid example can be seen in Figure \ref{demo}.We divide the tasks into training tasks and testing tasks with a ratio of 3:1. In detail, all instructional texts of 152 tasks are taken as training data, and texts of the remaining 50 tasks as testing data. 

\subsection{Baselines and Criteria}
\paragraph{Baselines}
We consider the following large language models to characterize the performance of our {\our} even with quite a low parameter amount.
\begin{enumerate}
    \item \textbf{Vanilla GPT2 family:}
    GPT2 is a transformers model pre-trained on a very large corpus of English data. Given the task, Vanilla GPT2 can generate corresponding instructional texts. \zf{In detail, we take the Chain-of-Thought(CoT) method to prompt GPT2. The task description and history actions are taken as input and GPT2 generates the next action.}
    We use the pretrained GPT2-small, GPT2-medium, GPT2-large and GPT2-xl with 124M, 355M, 774M, and 1.5B parameter amounts, respectively. 
    
    \item \textbf{fine-tuned GPT2 family:}
    We finetune the pre-trained GPT2 family using the same label texts as {\our} for fair comparison. \zf{In detail, we follow the CoT method to generate training data, where the texts of task description and history are labeled with the next action.}

    \item \textbf{ChatGPT(GPT-3.5):}
    ChatGPT is an artificial intelligence chatbot developed by OpenAI which owns a 175 billion parameter amount. As the latest milestone in interactive large language models, ChatGPT has exhibited human-level performance on various professional and academic benchmarks.
    \item \textbf{GPT4:}
    GPT4 is a multimodal large language model which owns a 1.8 trillion parameter amount. GPT4 is more creative and collaborative than ever before. Compared with previous models which include GPT-3.5, GPT-4 has better human-like conversations and provides more accurate results.
    \item \textbf{SayCanPay:}
    \base{SayCanPay is a model that combines the power of LLMs and heuristic planning by leveraging the world knowledge of LLMs and the principles of heuristic search \cite{hazra2023saycanpay}. In detail, the SayCanPay model leverages the world knowledge of Say model and train Can and Pay models using expert data. After that, it uses greedy search and beam search to generate actions. Thus, we re-train the can and pay models and then test using greedy search and beam search, separately.}
    % employs LLMs to generate actions (Say) guided by learnable domain knowledge, that evaluates actions' feasibility (Can) and long-term reward/payoff (Pay), and heuristic search to select the best sequence of actions.
    % \item  \textbf{Translated GPT2-xl (Huang model)}
    % Translated GPT2-xl (Huang model) is proposed by Huang et al. \cite{huang2022language}, which conditions on existing demonstrations and semantically translates the plans to admissible actions. 
\end{enumerate}

\paragraph{Evaluation Criteria}
Specifically, we follow Huang et al. \cite{huang2022language} to define the executability and correctness. In addition, we use BLEU and ROUGE from the NLP field to evaluate the reasonability of generated texts. \\
\textbf{Executability} measures whether all actions in an action plan satisfy the common-sense constraints of the environment. Specifically, each action (e.g. grab milk from the fridge) can be executed only if the corresponding pre-conditions (e.g. the fridge is open) are satisfied. \zf{One script is executable if and only if all preconditions of actions in the script can be satisfied. Then, we calculate the average proportion of executable scripts among all generated scripts for executability.}\\
\textbf{Correctness} evaluates whether one script can ensure the task completion. The complex task often contains one or more goal states that require to be satisfied, which is difficult to judge by automation metrics. Thus, we ask human evaluators to judge whether one script has the correct logic and completes the given task by their experiences. Then, we calculate the proportion of correct scripts among all generated scripts for correctness.

Unlike embodied environments like the OFFICEWORLD domain and Montezuma’s Revenge domain \cite{icarte2018using,mnih2015human}, the ambiguous nature of natural language task description makes it hard to find a perfect standard for the correctness of generated instructions. Referred to Huang et al. \cite{huang2022language}, we thus commit human evaluation to measure whether the generated instructions ensure the task completion from human common sense.
We randomly select 20 tasks in the testing datasets. Then we invite 50 human annotators to measure whether the generated instructional texts are correct. The correctness rate of generated texts is calculated by averaging the correctness ratio from different human annotators.\\
\textbf{BLEU} measures the similarity between generated texts and reference texts based on n-gram (Equation (\ref{bleu})) \cite{papineni2002bleu}. In detail, BLEU detects the common segments with different lengths between generated texts and reference texts. The 1-gram BLEU detects the same word and evaluates how many redundant words the generated instructions have.  The n-gram ($\geq1$) detects textual segments and evaluates how the generated actions submit to actions in label texts.
\begin{equation}
\label{bleu}
{BLEU}_{n}=\frac{\sum_{c \in {texts}} \sum_{n_{gram} \in c} {Count}_{clip}(n_{gram})}{\sum_{c^{\prime} \in {texts}} \sum_{n_{gram^{\prime}} \in c^{\prime}} {Count}\left(n_{{gram}^{\prime}}\right)}
\end{equation}
where $texts$ represents the sentences in generated instructional texts. ${Count}_{clip}(n_{gram})$ and ${Count}\left(n_{gram^{\prime}}\right)$ denote the number of each $n_{gram}$ and $n_{gram^{\prime}}$ from instructional texts appear in both and candidates, respectively.\\
\textbf{ROUGE} measures the similarity between generated texts and reference texts (Equation (\ref{rouge})) \cite{lin2004rouge}. Different from BLEU, it can consider the overlapping of n-gram, the longest common subsequence, and weights of different words.  
\begin{equation}
\label{rouge}
{ROUGE}_{n}=\frac{\sum_{c \in {texts}} \sum_{n_{gram} \in c} {Count}_{clip}(n_{gram})}{\sum_{c^{\prime} \in {labels}} \sum_{n_{gram^{\prime}} \in c^{\prime}} {Count}\left(n_{{gram}^{\prime}}\right)}
\end{equation}
where $labels$ represents the sentences in label texts. ${Count}_{clip}(n_{gram})$ and ${Count}\left(n_{gram^{\prime}}\right)$ denote the number of each $n_{gram}$ and $n_{gram^{\prime}}$ from instructional texts appear in both and only references, respectively.
% In the following experiments, we empirically evaluate {\our} and all baselines using BLEU and ROUGE referred to \cite{papineni2002bleu} and \cite{lin2004rouge}.
% \begin{equation}
% \label{bleu}
% {BLEU}_{n}=\frac{\sum_{c \in {texts}} \sum_{n_{gram} \in c} {Count}_{clip}(n_{gram})}{\sum_{c^{\prime} \in {texts}} \sum_{n_{gram^{\prime}} \in c^{\prime}} {Count}\left(n_{{gram}^{\prime}}\right)}
% \end{equation}
% where $texts$ represents the sentences in generated instructional texts. ${Count}_{clip}(n_{gram})$ and ${Count}\left(n_{gram^{\prime}}\right)$ denote the number of each $n_{gram}$ and $n_{gram^{\prime}}$ from instructional texts appear in both and candidates, respectively.
% \begin{equation}
% \label{rouge}
% {ROUGE}_{n}=\frac{\sum_{c \in {texts}} \sum_{n_{gram} \in c} {Count}_{clip}(n_{gram})}{\sum_{c^{\prime} \in {labels}} \sum_{n_{gram^{\prime}} \in c^{\prime}} {Count}\left(n_{{gram}^{\prime}}\right)}
% \end{equation}
% where $labels$ represents the sentences in label texts. ${Count}_{clip}(n_{gram})$ and ${Count}\left(n_{gram^{\prime}}\right)$ denote the number of each $n_{gram}$ and $n_{gram^{\prime}}$ from instructional texts appear in both and references, respectively.

% Then, we measure the executability and correctness of instructional texts. 
%不考虑最优性
\zf{In this paper, we do not consider the optimality of solutions, where the optimal solution of a task is defined as the shortest action sequence that can ensure the task completion. This is because it is challenging to generate the optimal solution even in AI Planning. Thus, the expert data we use are often inoptimal and may contain redundant actions. Limited by such training data,  it is hard to ensure the optimality of solutions. 
% In our experiments, we will focus on the executability and correctness of action sequences
} 
% We calculate the average proportion of executable texts among all generated texts for executability. 
% We also hire humans to judge the correctness and legibility of generated instructional texts referred to \cite{huang2022language}. Since human resource is expense and time-consuming, we randomly select 10 tasks in the testing datasets. Then we invite 50 human annotators to measure whether the generated instructional texts are correct. The correctness rate of generated texts is calculated by averaging the correctness ratio from different human annotators.

\subsection{Experimental Results}
Our experiments evaluated the text quality and practicability of generated instructional texts. We will examine our {\our} in the following aspects:
\begin{itemize}
    \item[1] We first evaluate {\our} on classic NLP metrics like BLEU and ROUGE, which measures the precision and recall of generated instructional texts with label texts and reflects the text quality and reasonability to some degree.
    \item[2] We then evaluate {\our} on executability and correctness of instructional texts. We judge the executability and correctness by the completion of the task and human, respectively.
    \item[3] Additionally, we evaluate the effects of language model guidance on {\our} by modifying training epochs and evaluating the generated instructions.
    \item[4] We adjust the weights of different heuristic values in graph search. We aim to see the impacts of heuristic values and find the optimal hyperparameter pair.
    % Our model can generate instructional texts with different lengths by setting the expected program length.
    \item[5] Shorter instructional texts often have higher executability, which may disturb the evaluation of text quality. To validate the robustness of our model, we finally evaluate the executability of generated instructional texts with varying program lengths.
    \item[6] Finally, we commit a case study by analyzing instructional texts of one specific task generated from {\our} and all baselines. 
\end{itemize}
\textbf{Text quality of generated instructional texts}

The BLEU and ROUGE score of generated instructional texts with label texts can reflect the text quality to some degree. \joey{In detail, we leverage ground truth instructional texts provided by \cite{huang2022language}.} We then compare BELU and ROUGE score among {\our} and all fine-tuned baselines. Table (\ref{brtab}) shows \joey{BLEU} and ROUGE results from different language models. From Table (\ref{brtab}), we can observe that our {\our} outperforms all baselines and achieves state-of-the-art performance on both BLEU and ROUGE score with the least parameter amount, which demonstrates the reasonability and submissiveness to label texts of our generated instructional texts. 
\begin{table}[!ht]
\caption{The comparison of BLEU and ROUGE score.}
\centering
\begin{tabular}{cccc}
\toprule
Model & Parameter& BLEU & ROUGE\\
\midrule
{\our} & \textbf{35.3M} & \textbf{0.40} & \textbf{0.26} \\ %0.55428
% {\our}-w/o Logical-Graph & 35.3M & 0 & 0\\
% Vanilla GPT2-small & 124M & 0\\ %0.34572
% Vanilla GPT2-medium & 355M & 0\\
% Vanilla GPT2-large & 774M & 113.2\\ %0.39429
% Vanilla GPT2-XL & 1.5B & 0\\
Tuned GPT2-small & 124M & 0.05 & 0.17\\
Tuned GPT2-medium & 355M & 0.05 & 0.15\\
Tuned GPT2-large & 774M & 0.05 & 0.16\\
Tuned GPT2-xl & 1.5B & 0.04 & 0.17\\
SayCanPay (Beam) & 218M & 0.04 & 0.15 \\
SayCanPay (Greedy) & 218M & 0.03 & 0.16 \\
% Translated GPT3(Huang model) & 0 & 0\\
% GPT-neo & 1.3B & 127.5 \\ % 0.16283
\bottomrule
\end{tabular}
\label{brtab}
\end{table}
\\
\\
\\
\textbf{Executability and correctness of generated instructional texts}

\begin{figure}[!t]
    \centering
    \includegraphics[width=0.5\textwidth]{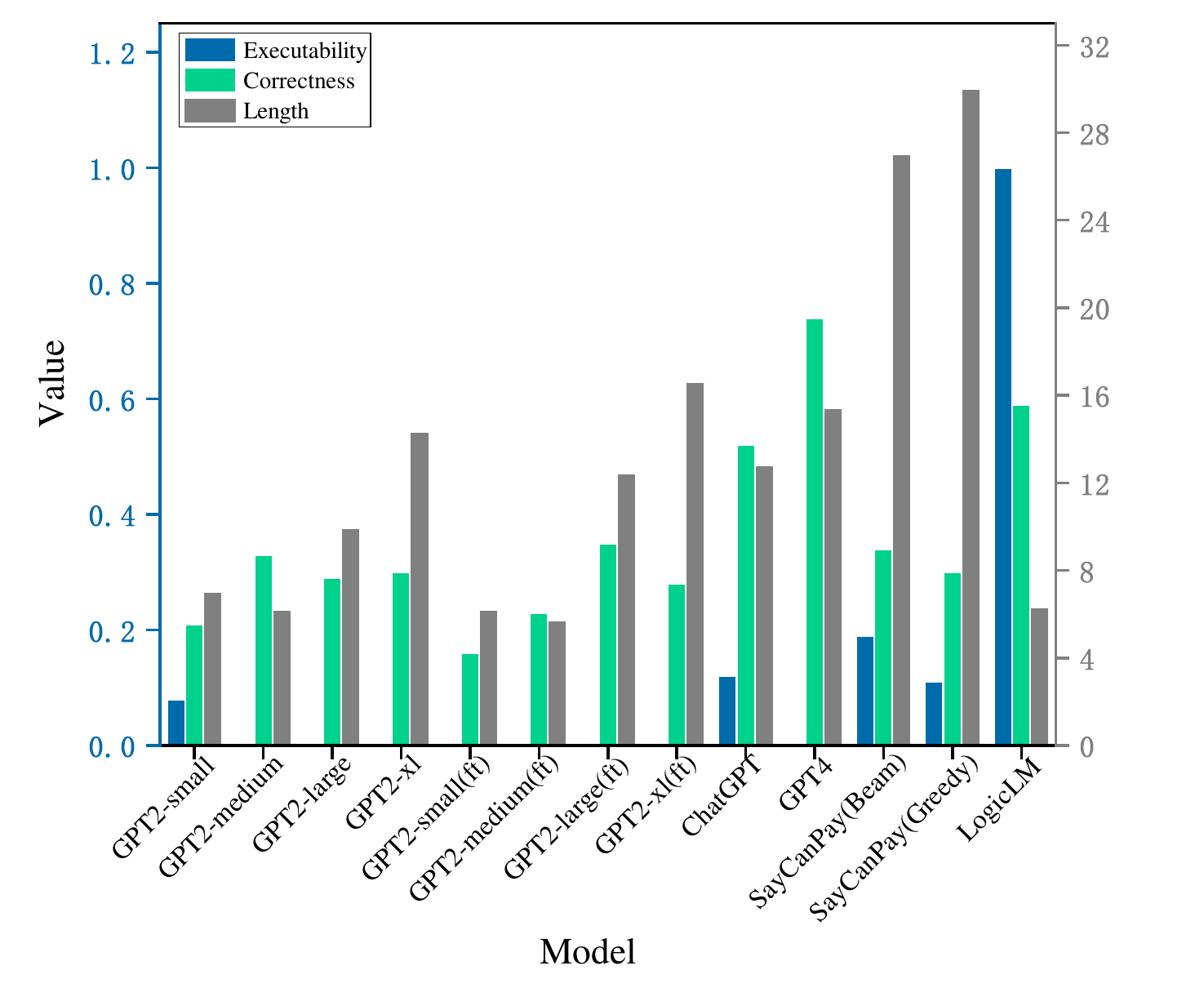}
    \caption{Quality measure of instructional texts.}
%% 两边大小差距太多，看起来有点不好看
    \label{exe_cor}
\end{figure}

The final goal of our {\our} is to complete tasks in sophisticated open environments. We thus compare the executability and correctness of generated instructional texts. In this experiment, executability indicates each action from instructional texts is effective and feasible. We judge correctness by humans and it implies instructional texts can ensure the completion of tasks. Specifically, we use the same questionnaires as Huang et al. \cite{huang2022language}, but we take the average correctness of evaluated tasks over all human evaluators for each model. The results together with the average program lengths are shown in Figure \ref{exe_cor}. Considering the great expenses of human evaluation, we randomly pick 20 tasks from evaluated tasks for human evaluation. In Figure \ref{exe_cor}, we see that our model performs better than all comparison models including ChatGPT. It is not surprising because our {\our} generates instructional texts based on domain-specific logical graph. Also, we note that GPT2-small has the highest executability among all models except our {\our}. The reason we believe is that the program length of instructional texts from GPT2-small is quite small than all other models and shorter instructional texts tend to have higher executability. In comparison, the program length of instructional texts from our {\our} is between those of GPT2-medium and GPT2-large but still performs better than almost all comparison models. Moreover, it can be observed that \zf{the fine-tuned GPT2 generates unexecutable instructional texts even if the original GPT2 small produced some executable output. We attribute this to that the scarce training data and finetuning LLMs may damage the safety strategies of LLMs.} Note that GPT4 gets higher correctness but lower executability compared to our {\our}. The higher correctness is not surprising because GPT4 has 1 trillion parameters and the enormous training corpus accidentally contains the required world knowledge known by humans. However, GPT4 may lack the required strict domain logic, resulting in lower executability. Moreover, our {\our} surpasses both the beam search and greedy search of SayCanPay. The reason we believe is that SayCanPay essentially internalizes domain logic using language models, which calls for sufficient expert data. Intuitively, our {\our} works better for generating instructional texts when domain-specific training data is scarce.\\

\noindent\textbf{Influence measure of language model guidance}

\begin{figure}[!ht]
\centering
\subfigure[BLEU and ROUGE]{
    \begin{minipage}[b]{0.22\textwidth}
    \includegraphics[width=\textwidth]{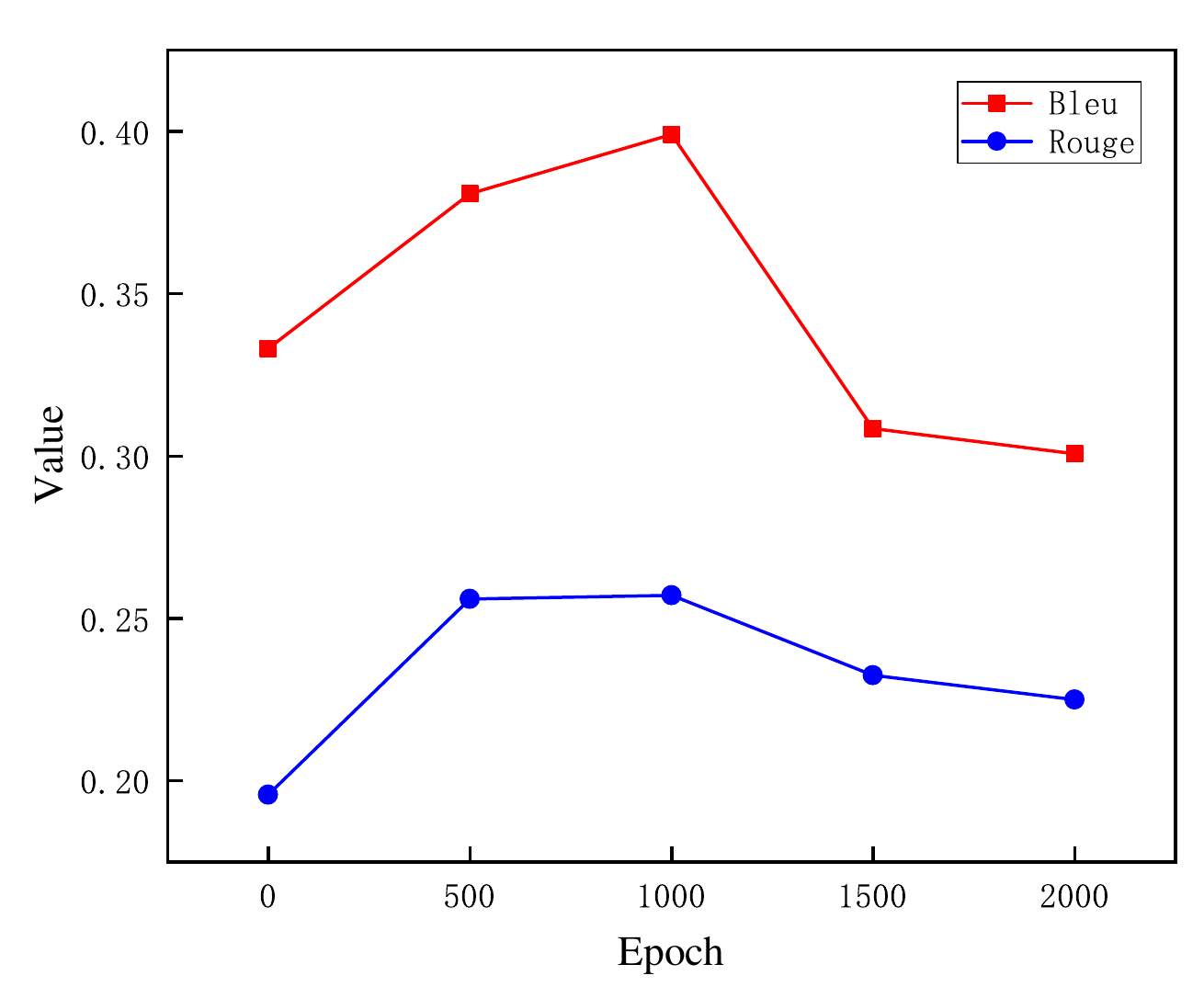}
    \label{br}
    \end{minipage}
}
\subfigure[Executability]{
  \begin{minipage}[b]{0.22\textwidth}
    \includegraphics[width=\textwidth]{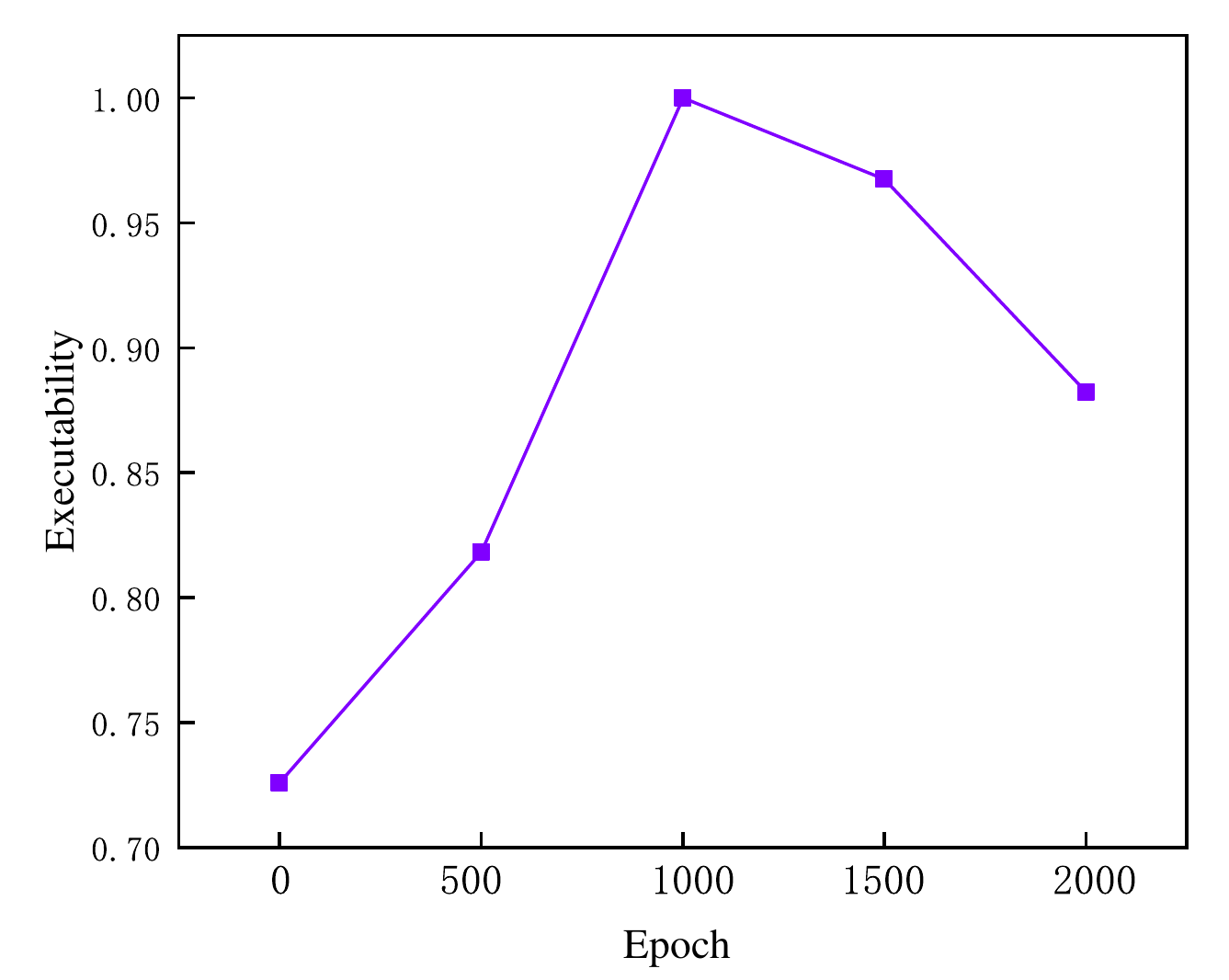}
    \label{exe}
  \end{minipage}
}
\caption{The text quality trends with varying epochs}
\label{brexe}
\end{figure}

To evaluate the effects of language model guidance on {\our},  we vary the training epochs of language models for different quality of language model guidance. We compare generated texts from {\our} on evaluated tasks with varying quality of language model guidance. As shown in Figure \ref{br}, both the BLEU score and ROUGE score rise first and then decline with the improvement of training epochs. The same trend of executability with varying quality of language models is also depicted in Figure \ref{exe}. The early rising trend shows that our language model guidance influences {\our} and higher training epochs indicate better guidance. Although higher epochs indicate better language model guidance, we should note that overlarge training epochs result in over-fitting, which illustrates the latter declining trend. \zf{Note that the over-fitting is unavoidable since our lightweight {\our} applies to domains with few expert samples. Thus, we divide the expert data into training data and validation by the ratio 8:2 and we leverage the effects of validation samples to break off training in advance.} In Figure \ref{exe}, we also see that the executability is still high at epoch 0. This is not surprising because our {\our} is also guided by bayes heuristic value, which contains domain logic.\\
% indicates language model can help select logical actions and generate more logical instructional texts. This both shows that our language model guidance influences {\our} and better guidance indicates better instructional texts.   

% \begin{figure}[!ht]
%     \centering
%     \includegraphics[width=8cm]{fig/br_epoch.pdf}
%     \caption{Bleu and Rouge trends with varying epochs}
% %% 两边大小差距太多，看起来有点不好看
%     \label{br_epoch}
% \end{figure}
% \begin{figure}[!ht]
%     \centering
%     \includegraphics[width=8cm]{fig/exe_epoch.pdf}
%     \caption{Executability trends with varying epochs}
% %% 两边大小差距太多，看起来有点不好看
%     \label{exe_epoch}
% \end{figure}
\noindent\textbf{\joey{Executability} with respect to various hyper-parameters}

\begin{figure}[!ht]
    \centering
    \includegraphics[width=0.5\textwidth]{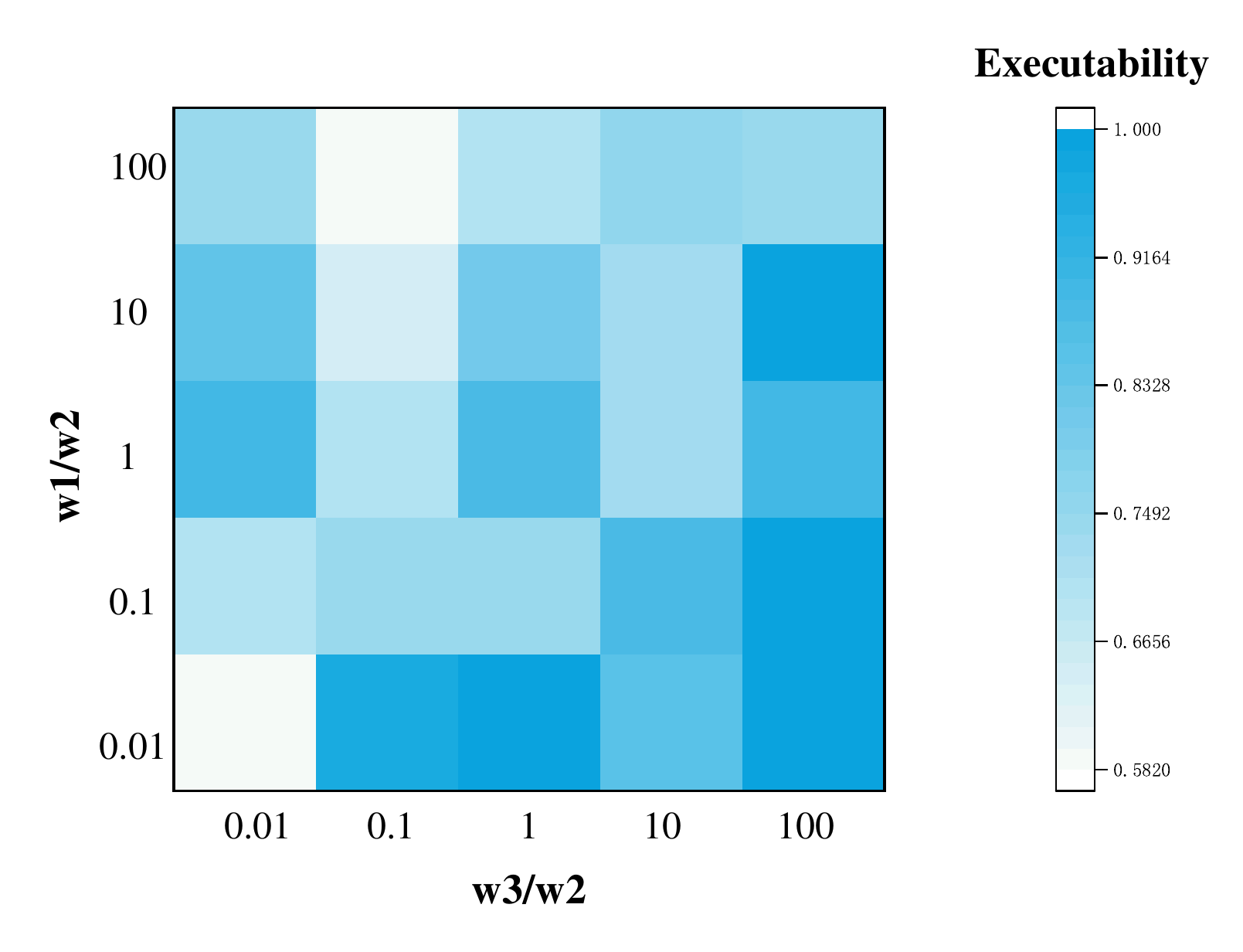}
    \caption{The heatmap of executability with different hyper-parameters}
%% 两边大小差距太多，看起来有点不好看
    \label{hyper}
\end{figure}

To see the impact of different hyper-parameters, we evaluate the executability of generated instructional texts on evaluated tasks with respect to different weights of heuristic values(i.e., $w_1$, $w_2$, and $w_3$ in Equation (\ref{totalheu})). The detailed heatmap about how hyper-parameters influence executability is depicted in Figure \ref{hyper}. \joey{We can fix the value of one axis and study how executability changes with another axis value.} In most cases, we can see that executability of generated instructional texts drops with $w_1$ increasing. A big value of $w_1$ means to focus on generated instructional text length and thus ignore text logic. When $w_3:w_2 = 100:1$, the executability of generated texts is the highest for the same $w_1:w_2$. The reason we believe is that the heuristic value $h_{lm}(.)$ is much smaller than $h_{bayes}(.)$. The $h_{lm}(.)$ can have considerable effects as $h_{bayes}(.)$ only when the weight of $h_{lm}(.)$ is much higher than that of $h_{bayes}(.)$.\\

\noindent\textbf{Controllable program length study}

\begin{figure}[!t]
    \centering
    \includegraphics[width=0.5\textwidth]{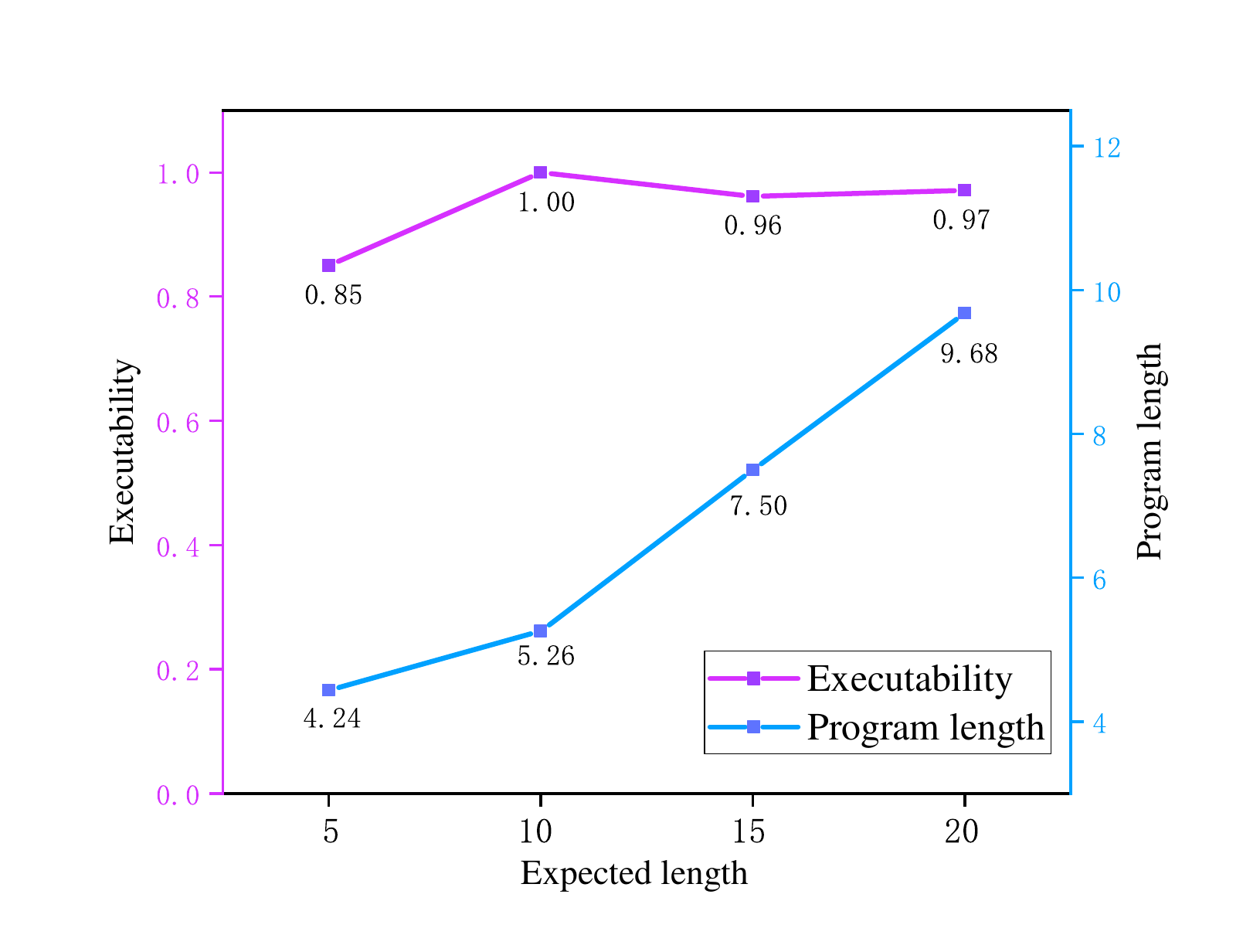}
    \caption{The robustness of our model for varying expected program length}
%% 两边大小差距太多，看起来有点不好看
    \label{len}
\end{figure}

We next evaluate controllable program length ability of our model and the performance with regard to the answer length. Since the graph search uses the expected program length as heuristic guidance, the program length of generated instructions is human controllable. In Figure \ref{len}, we see that the program length (blue lines) truly changes with different program length. However, the generated program length change is not obvious. The small change range is because the training instances for constructing the logical graph have short program lengths and there exist no long paths in the logical graph. To validate the robustness of our {\our}, we also compare executability of generated program length as the expected program length changes. From purple lines in Figure \ref{len}, we can observe that the executability soars as the program length increases. In the future, we hope to construct logical graph with longer paths and further explore the program length robustness of our {\our}.
\begin{figure}[!t]
    \centering
    \includegraphics[width=0.5\textwidth]{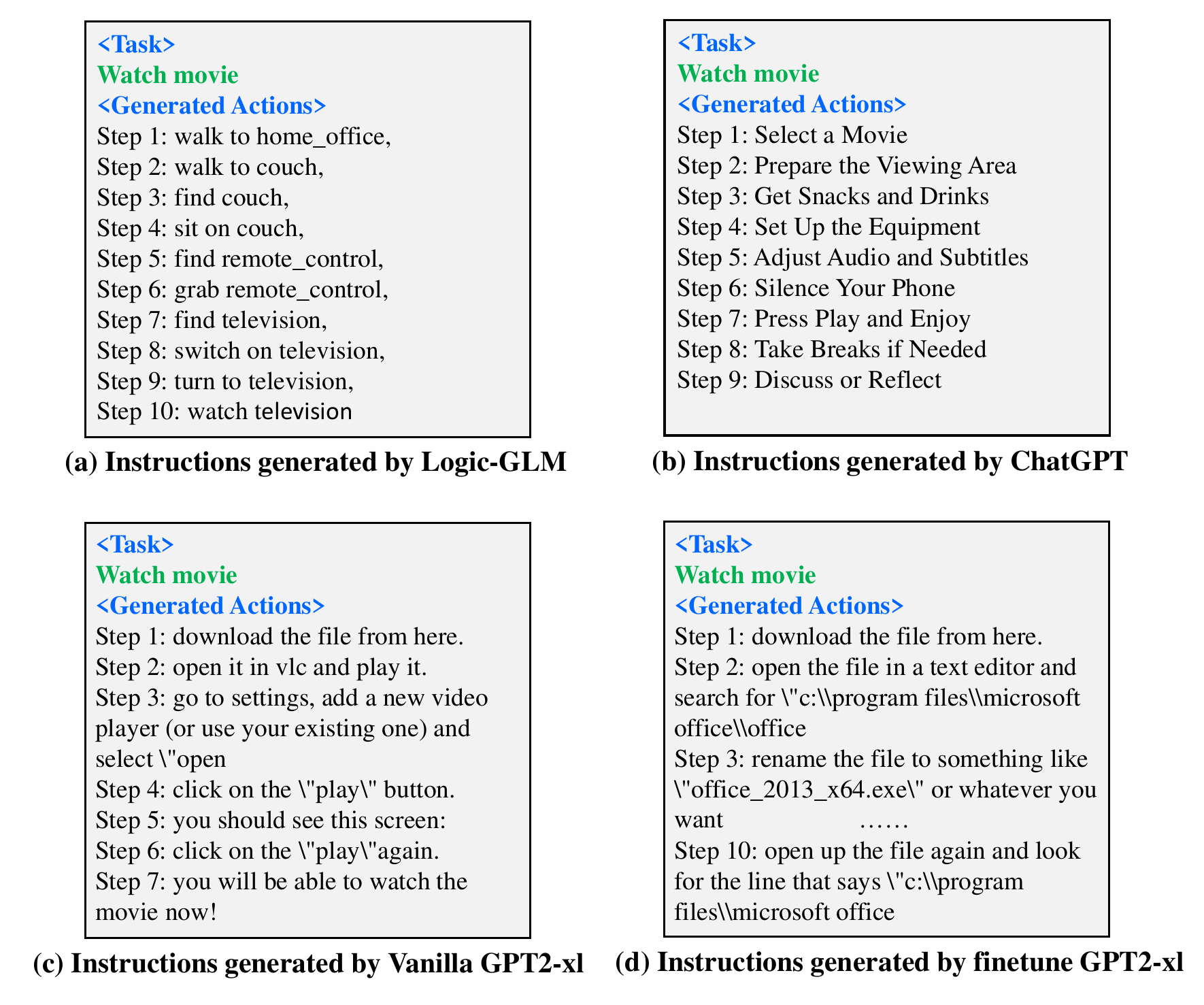}
    \caption{A case study on task "Watch movie"}
%% 两边大小差距太多，看起来有点不好看
    \label{case}
\end{figure}

\noindent\textbf{Case study on task "Watch movie"}

To vividly demonstrate the difference between generated instructional texts from different language models, we design an experiment to analyze the outputs of {\our} and other language models. In detail, we choose ChatGPT and both Vanilla and fine-tuned GPT2-xl which has the higher parameter amount among GPT2 family. We sample the task "Watch movie" and demonstrate instructional texts from different models in Figure \ref{case}. It is intriguing to note that our model can handle the task well. 

From Figure \ref{case}, we can observe that both our {\our} and ChatGPT can generate reasonable instructional texts relevant to the given task, which indicates that our {\our} can achieve comparative performance as ChatGPT with the guidance of domain-specific logical graph. It is obvious that the instructional texts generated from our {\our} are also simple and have higher consistency between sentences. Moreover, We can observe that both Vanilla GPT2-xl and Finetune GPT2-xl generate much more sophisticated texts which lack consistency between sentences and include many non-critical objects in the specific task.

\section{Related Work}
\label{rel}
Along with rapid advances in language models, there has been a greater interest in uncovering the reasoning abilities of such models \cite{huang2022towards,pan2023logic, qiao2022reasoning}. The research on reasoning includes two parts. A significant part of this progress is to extract logical rules from texts using language models. For example, Cresswell et al. \cite{plantext1,plantext2} use extracted rules to construct domain models for planning.  Also, Huang et al. \cite{huang2022language} directly uses the pre-trained large language models for generating logical instructional texts. Li claims that large language models possess simple reasoning abilities \cite{li2021implicit}. With respect to AI planning, there have been lots of works on integrating AI planning with natural language processing (c.f. \cite{DBLP:journals/corr/abs-2202-07138, DBLP:journals/corr/abs-2202-08373,DBLP:conf/ijcai/FengZK18}). Large language models, however, fail to reason in some special or sophisticated scenes \cite{valmeekam2022large}. In this work, we aim to combine language models with AI planning for augmenting the logical reasoning abilities.

A variety of works have been studied in capturing domain logic and augmenting the reasoning abilities of models \cite{kokel2021dynamic,badreddine2022logic,dash2022review}. Song et al. \cite{graph2text2,graph2text3} represent domain logic with knowledge graph and then guide the text generation. Also, Zhang et al. \cite{zhang2022persona} use a graph to represent relationships between events and personas for persona-guided text generation. Similarly, our approach leverages both AI planning and graph structure for quick domain knowledge capturing. However, the graph structure and language models in our approach guides each other to make logical policies in specific environments.

\section{Conclusions}
\label{con}
In this paper, we propose a novel approach {\our} to generate logical skeletons which can be infused into language models with the smallest training expenses. In detail, the language model and logical graph guide each other to internalize the domain logic. In addition, the search process generates skeletons according to heuristic values and indirectly explains the policy of language models to generate texts. \zf{We show that our {\our} can be used in novel scenes of the same domain. The generalizability of the lightweight model towards scenes of different domains deserves further exploration. }In the future, it would be interesting to investigate the integration of planning model learning \cite{DBLP:conf/aips/ZhuoYPL11,DBLP:conf/aaai/Zhuo15} and plan recognition techniques \cite{DBLP:conf/aaai/Zhuo17, DBLP:journals/tist/Zhuo19,DBLP:journals/tist/ZhuoZKT20} for interpretably generating instructions of high-quality.

%%%%%%%%%%%%%%%%%%%%%%%%%%%%%%%%%%%%%%%%%%%%%%%%%%%%%%%%%%%%%%%%%%%%%%%%

%%% Use this environment to include acknowledgements (optional).
%%% This will be omitted in doubleblind mode.

\begin{ack}
This research was funded by the National Natural Science Foundation of China (Grant No. 62076263).
\end{ack}
% By using the \texttt{ack} environment to insert your (optional) 
% acknowledgements, you can ensure that the text is suppressed whenever 
% you use the \texttt{doubleblind} option. In the final version, 
% acknowledgements may be included on the extra page intended for references.
% \end{ack}

%%%%%%%%%%%%%%%%%%%%%%%%%%%%%%%%%%%%%%%%%%%%%%%%%%%%%%%%%%%%%%%%%%%%%%%%

%%% Use this command to include your bibliography file.
%\clearpage
% \bibliographystyle{plain}
\bibliography{mybibfile}

\end{document}